\documentclass[letterpaper, 10pt, conference, dvipsnames]{ieeeconf}
\IEEEoverridecommandlockouts

\usepackage[pdftex]{graphicx}
\usepackage{multicol}
\usepackage{amsmath, amsfonts, amssymb}
\usepackage{iasbib}
\usepackage{xspace}
\usepackage{latexsym}
\usepackage{url}
\usepackage{listings}
\usepackage{iaslistings}
\usepackage{float}
\usepackage{tabularx}
\usepackage{tabulary}
\usepackage{tikz}
\usepackage{pgfplots}
\usepackage{fancybox}
\usepackage{flushend}
\usepackage{lipsum}
\usepackage[ruled,lined]{algorithm2e}
\usepackage{hyphenat} 
\usepackage{xcolor}
\lstdefinestyle{Prolog} {language=Prolog,
                         lineskip=-0.3ex,
                         fontadjust=true,
                         basicstyle={\footnotesize\nopagebreak[4]},
                         commentstyle=\footnotesize,
                         keywordstyle=\footnotesize,
                         showstringspaces=false,
                         showspaces=false,
                         showtabs=false,
                         moredelim=**[is][\bf]{@}{@},
                         moredelim=**[is][\it]{~}{~}
                         }
\lstdefinestyle{Lisp}   {language=Lisp,
                         lineskip=-0.5ex,
                         fontadjust=true,
                         basicstyle={\footnotesize \nopagebreak[4]},
                         commentstyle=\footnotesize,
                         keywordstyle=\footnotesize,
                         moredelim=**[is][\it\bf\color{BrickRed}]{<}{<},
                         moredelim=**[is][\it\bf\color{Violet}]{&}{&},
                         moredelim=**[is][\it\bf\color{OliveGreen}]{^}{^},
                         moredelim=**[is][\it\bf\color{BurntOrange}]{!}{!},
                         moredelim=**[is][\bf]{@}{@},
                         moredelim=**[is][\it]{~}{~}
                         }

\usetikzlibrary{matrix}
\usepgfplotslibrary{groupplots}
\pgfplotsset{compat=newest}

\newcommand{\cram}{\textsc{CRAM}\xspace}
\newcommand{\robosherlock}{\textsc{RoboSherlock}\xspace}
\newcommand{\knowrob}{\textsc{KnowRob}\xspace}

\newcommand{\neems}{\textsc{NEEMs}\xspace}
\newcommand{\urobosim}{\textsc{URoboSim}\xspace}
\newcommand{\bulletworld}{Bullet\xspace}
\newcommand{\giskard}{Giskard\xspace}

\graphicspath{{../../../img/}}

\title{URoboSim --- \\ An Episodic Simulation Framework for Prospective
  Reasoning in Robotic Agents}

\author{Michael Neumann, Sebastian Koralewski and Michael Beetz*\\
        {\{neumann.michael, seba, beetz\}@cs.uni-bremen.de}
\thanks{*The authors are with the Institute for Artificial Intelligence,
University of Bremen, Germany.}}

\begin{document}

This work has been submitted to the IEEE for possible
publication. Copyright may be transferred without notice,
after which this version may no longer be accessible.
\clearpage

\maketitle

\begin{abstract}

  Anticipating what might happen as a result of an action is an essential ability 
  humans have in order to perform tasks effectively. On the other hand, robots capabilities in this regard are quite lacking.
  While machine learning is used to increase the ability of prospection it is still limiting for novel situations. A possibility 
  to improve the prospection ability of robots is through simulation of imagined motions and the physical results of these actions.
  Therefore, we present URoboSim, a robot simulator that allows robots to perform tasks as mental simulation
  before performing this task in reality.
  We show the capabilities of URoboSim in form of mental simulations,
  generating data for machine learning and the usage as belief state for a real robot.

\end{abstract}


\section{Introduction}
\label{sec:intro}

A distinctive aspect in the human mastery of manipulation tasks is the
leverage of prospection to go beyond the immediate sensory-motor
experience and make better informed decisions about their intended
course of action. When a human pours pancake mix onto a pancake maker
she can parameterize and adapt the movements such that at the end the
pancake mix forms a circular pancake of a certain size. She can also
immediately answer questions regarding to what will happen if she
holds the pancake mix too high, tilts it too fast, if the pancake is
too thick, or too thin. These prospective capabilities enable humans
to quickly predict and diagnose the causes of unwanted side effects
and adapt the movements to forestall them.

Prospection, the ability to represent what might happen in
the future, enables humans to carry out tasks effectively by
anticipating and taking into account the predicted effects of their
actions. The prospective capabilities of the human brain are varied
and very sophisticated including simulation, prediction, intention
formation, and planning. Within this collection of prospective
capabilities (episodic) simulation, also called motor imagery, plays a
central role. Simulation (motor imagery) is the process of imagining
an action without executing the movements involved. Hesslow in his
simulation theory of cognition even argues that a substantial subset
of human cognitive capabilities are based on mental simulation.

Compared to the rich and sophisticated prospective capabilities of
humans those of today's robotic agents are still quite rudimentary. In
particular, a simulation mechanism that can generate detailed
representations of imagined movements and their physical effects and
allows for open question answering about the simulated episode is still
missing.

Several researchers have proposed to learn such simulation
capabilities through machine learning methods in particular using deep
artificial neural networks \cite{paxton}. While these approaches are
significant advancements of prospective capability they are still too
limited to enable the mastery of the variety of object manipulation
tasks that humans accomplish in their everyday activities. One
limitation is that these simulators are learned only from experience
data which questions their applicability to novel situations and
movements for which mental simulation would certainly be particularly
important \cite{Davis2014TheSA}. Second, the representation of
environment states, the evolution of states, and the causal relation
between movements and their physical effects is only implicitly
represented in the weights of the artificial neurons and not
accessible for other reasoning methods. 

In this paper, we propose to leverage the powerful simulation and
rendering mechanisms of today's virtual reality technologies in order
to develop more powerful mental simulation capabilities. The advantage
of these simulation mechanisms over ones learned from data are, that
physics simulators hold explicit data structures for the
representation of states and the evolution of states that can be
leveraged for predictive reasoning.  The simulation algorithms
also implement basic laws of nature that have a broad scope of
validity such as Newton physics or the laws of fluid motions and can
therefore make informative predictions outside the scope of prior
experience.

Of course, there is the common objection that ``simulations are doomed
to succeed'', that is that simulations will not be able to
realistically cover all aspects of what should be simulated. In
practice, it is often not the simulation algorithm but rather unknown
physical parameters that prevent accurate prediction. When pouring
pancake mix the pouring agent typically does not know the viscosity
and surface friction necessary for simulation. This problem occurs
independently of what simulation approach is chosen. For certain
tasks, however, this knowledge is not required. For example, you can
learn to pour pancake in terms of how the pancake changes its shape
during pouring. This representation is independent of the individual
physics parameters and therefore can even be learned with the
parameter values being wrong. Looking at human manipulation episodes
we see how often and immediately humans respond to action failures
suggesting that prospection is an information resource about possible
futures rather than a predictor for the right future. 

The novel contribution of this paper is URoboSim, a robot simulator that
allows us to:
\begin{itemize}
\item represent an artificial world in simulation as well as symbolic knowledge base
 \item execute plans from the real robot in the artificial world
\end{itemize}

We evaluate our approach by using URoboSim to improve the performance of a fetch and deliver action.

\section{Transition from Real- to Artificial World}
\label{sec:perception-action}
\begin{figure}[htb]
  \centering
  \includegraphics[width=\columnwidth]{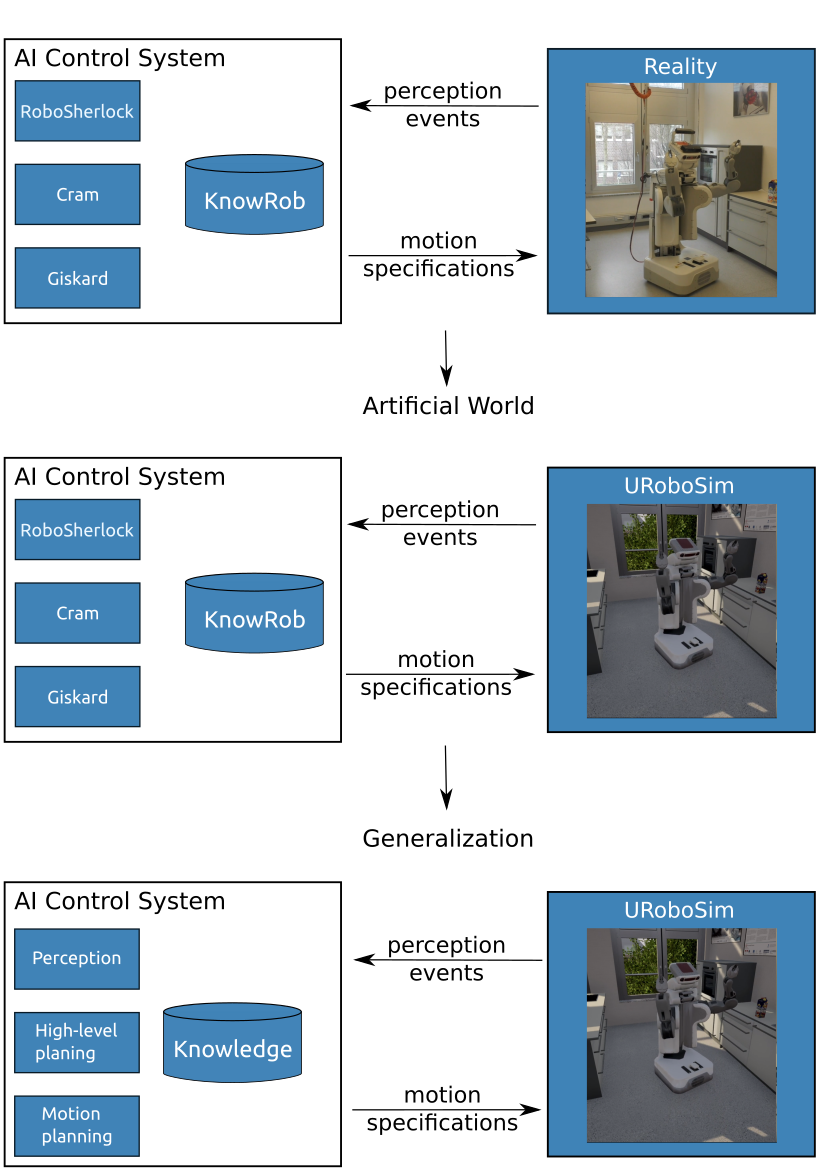}
  \caption{Perception-action loop of the real robot (top), the artificial world (middle) and a generalized approach (bottom).}
  \label{fig:perception-action}
\end{figure}

For robots to prospect about the future or generate data for learning problems,
we have to be able to execute actions in simulation in the same way as on the real robot.
This means the simulation must be able to replace the real robot and world in the perception-action loop of the robot.

In Figure \ref{fig:perception-action}, we depict the transition of the perception-action loop from the real world (top), to an artificial world (middle) and finally to a generalized
artificial world (bottom) that allows the usage of a custom AI control system.

The perception-action loop of our real robot consists of the
world with the robot and an AI control framework. The AI control system communicates with the robot and the world through perception events.
This includes image data as well as force dynamic events such as collision or grasping events. 
The AI control system consists of the perception framework \robosherlock \cite{beetz2015robosherlock},
the high-level planning framework \cram \cite{cram}, the robot motion control system \giskard~\cite{fang2016learning} and the knowledge processing system \knowrob.
\robosherlock uses the image data from the robot as well as knowledge from \knowrob~\cite{beetz18knowrob} to classify perceived objects
and build the scene graph. Given a task, \cram uses the knowledge from \knowrob and the scene graph to plan and start the next robot action. \cram~itself uses a simulator
based on bullet \cite{bullet} in order to prospect about the result of action \cite{kazhoyan19projection}. The simulation is simplified in order to be fast. Among others, it tests if 
a target robot position is valid or collides with the environment. It does not test the continuous transition of states during actions such as opening or closing drawers.
The simulator also does not take motion control or perception into account.
If the robot has to move its base or joints, for example for reaching actions, \giskard is used to plan the motion of the base and joints
so that collisions are avoided.

In order to transition to the mental simulation mechanism, the artificial world is created through \urobosim. This artificial world is
interchangeable with the real world. For this, the artificial world contains the body of the robot, the objects that it is to perceive and
manipulate and the environment in which it is to operate. The artificial world and the robot
interact in two ways. First, given the camera specifications of the
robot body and the 6D pose of the camera in the environment the
artificial world renders an image from the current scene when the
robot triggers the camera to capture an image. Second, the motion
specification computed by the control program is simulated in the
environment in order to compute the physical effects and the sensor
events caused by the respective movement.

The main data structure of the mental simulation mechanism is the
\emph{scene graph} \cite{yang2018graph} of the artificial
world implemented through a virtual reality engine. The scene graph
provides the graphics system with a data structure that contains all
the information needed to render images of the 3D scene. The scene
graph specifies and organizes the static contents of a 3D scene as
well as the dynamic events and mechanisms (e.g. articulation models)
that can modify the 3D scene. The data structure is highly optimized
and facilitated by very fast and efficient algorithms for storing,
managing very complex and detailed scenes, for retrieving information
from it, and for rendering and simulation. A scene graph is an
abstraction that provides an interface between an application
developer and a rendering system.

The scene graph is a tree-like data structure where the nodes
represent subscenes and the child relations from a node \emph{p} to
the nodes \emph{c$_1$, \ldots, c$_n$} state that the subscene \emph{p}
contains all subscenes \emph{c$_1$, \ldots, c$_n$}.  The leaf nodes,
called shape nodes, represent the geometry, appearance, lighting,
texture, surface attributes, and material of the AW entity. The bodies
of the entities are typically CAD models or meshes. The inner nodes
hold a group or composition of children nodes that might be
constrained by physical relations such as attachment, support, and
articulation. Also, children nodes might represent the subscene at
different levels of detail based on the distance to the viewer.


\section{URoboSim}
\label{sec:urobosim}

URoboSim is a robot simulator implemented using
Unreal Engine 4. There are two unique features of URoboSim. The first one is
that the artificial world, in addition to the representation in simulation, is
represented by a symbolic knowledge base. This symbolic knowledge base
enables users of the artificial world to ask semantic questions about
the environment such as show me the place where perishable items are
stored or show me all containers in the environment and how they can
be opened (that is the articulation model of the container). The second
difference is that the activity in an episode is automatically
recorded and symbolically represented as an episodic
memory.
This episodic memory enables the user of the simulation
mechanism to ask queries about what the robot did, how it did it, why,
what it has seen when executing an action, how it moved to pick up an
object, and so on. The recording of the episodic memories enables open
question answering about an episode and the generation of supervised
learning problems.
We call those episodic memories - narrative enabled episodic memories (\neems).                     
A detailed overview of the representation about \neems~is given in \cite{neem:2020} 

Another notable feature of URoboSim is that it enables
to simulate the control programs that also run on the real robot. This
requires that the simulator also simulates the low-level data
structures and control abstractions of the physical control system,
including the kinematic chain of the robot and  joint motions.
In particular, the simulator provides the ROS low-level
control interface to the user for programming robot applications.
Making use of the photorealism in Unreal Engine 4, URoboSim enables the
perception to run object classification on image data generated from the simulation.
URoboSim was developed as a plugin for Unreal Engine 4.

Unreal Engine 4 itself, provides us with photorealism necessary for the perception and physics to
simulate interaction with the environment. Furthermore, by adding plugins to Unreal Engine 4
new functionalities can be added. For instance, if we want to simulate human-robot interaction, we could add UIAIAvatar \cite{uiaiavatar} in addition to \urobosim. UIAIAvatar 
provides functionalities to simulating humans.

URoboSim is implemented in a modular fashion. This means 
functionalities are enabled through adding components to the robot.
The robot itself can be loaded from SDF, an XML based robot description
developed by Gazebo \cite{gazebo}. The robot itself is only a model without additional functionalities.
By adding components, functionalities can be added. Those components include sensors such as LiDAR sensors
or cameras as well as different controllers

The simulator provides different controller among others:
\begin{itemize}
\item BaseController: This controller enables the robot to move in a holonomic way. This means the robot can move linearly in the x-y plane while rotating around the z axis. The linear motion is independent of the rotation.
\item JointController: Enables motion of the joint. It is possible to set the desired joint position directly or to follow joint trajectories. It is possible to move the robot kinematic or dynamic.
\item PR2HeadController: Moves the desired frame to point in the desired direction. This is specifically implemented for the PR2.
\item PR2GripperController: Moves the gripper to the desired position. This is specifically implemented for the PR2.
\end{itemize}

In order to communicate with the different parts of the perception-action loop over ROS, interfaces are implemented. The communication to ROS is realized over a
websocked using rosbridge\_suite \cite{rosbridge} and UROSBridge \cite{urosbridge}. It is possible to use topics, services
and actions to communication with the simulation. For the perception
framework, our simulator is able to publish RGB, depth and segmented
images. For the motion planing system, we publish the joint state and
odometry. Furthermore, we are publishing data of laserscans that
measure the distance to the surrounding.

\section{Applications}

Because our simulator has physics as well as photorealism we can use it in more advanced ways compared to other simulators such as bullet.
P. Mania and M. Beetz in \cite{mania19scenarios} already showed that we can use the photorealism in Unreal Engine 4 to generate
data for image classification. Using the possibility to use image data from simulation for perception, we closed the perception-action loop
using URoboSim. 

The simulator is currently used in three different ways. Firstly as mental simulations of real actions and secondly to generate
data for machine learning problems and thirdly as belief state of the robot.

\subsection{URoboSim as mental simulation}

Performing the action using URoboSim before executing the action on the real robot
allows us to answer different questions such as, is an object in the expected location visible from the specified robot position, if not from where would it be visible.
Another relevant question is what happens if a grasped object is released. By default, the robot does not know what happens to the object after the release.
While it is possible to add mechanisms to infer that the object is for example on the table if it's released above the table, this approach is limiting novel situations.
A more valid approach is to make use of physics engines to simulate the effect of releasing the grasped object. Although not all physical effects are taken into account,
it provides a good enough prediction to find it with the perception.

\subsection{Data Generation and Learning}
\label{sec:application-learning}
For the high-level planner inferring the correct plan parameters is a tremendous challenge.
Especially, when the agent needs to interact in a dynamic environment such as a kitchen.
To improve the high-level planner's inference performance, the planer can utilize statistical models created by learning from \neems.

In \cite{koralewski2019self} we transformed high-level plans into specialized plans using the experience from a robotic agent.
In general, \cram~itself does the parameter search by applying brute-force and heuristics.
In our previous work, we generated statistical models that reduce the search space and weight the remaining parameters by their success in the past.
To determine the success probability the models learn e.g.\ the best position for the robot to grasp the object with the best grasp configuration based on the current object pose.
For \cite{koralewski2019self} we acquired the \neems~by letting the robot perform pick-and-place experiments only in \bulletworld.
The resulting statistical models were then successfully transferred to the real robot.

Using simulators, such as \bulletworld~and \urobosim, have the advantage that experiments can be repeated very fast. 
There is no time required to reset the experimental setup. 
This allows us to generate much data in a short time.
However, we realized that once we started to move from simple pick-and-place experiments to more complex scenarios like setting up a breakfast table, the models created from \bulletworld~were starting to fail in the real world.
To name a specific scenario, our agent learned a distribution to grasp a milk box from the fridge door which showed promising results applied in the \bulletworld~simulation but failed completely in the real world.
The explanation for that behavior is that we do not have the same perception-action loop in \bulletworld~as in \urobosim. 
The ground truth about the world state in our \bulletworld~simulation causes the behavior that our agent can grasp the milk from the fridge door without the milk being in the robot's field of vision. 
However, in the real world, we implemented a safety mechanism which requires that the agent needs to perceive the object again before grasping it.
Applying the learned model from the \bulletworld~world causes the robot to position itself that it cannot perceive the milk again before grasping it.
To cope with that scenario we were required to generate statistical models from \urobosim.
A detailed description of how those models are generated, we will give in Section \ref{sec:ex}.


\subsection{Belief State}

\label{sec:belief-state}
\begin{figure}[htb]
  \centering
  \includegraphics[width=\columnwidth]{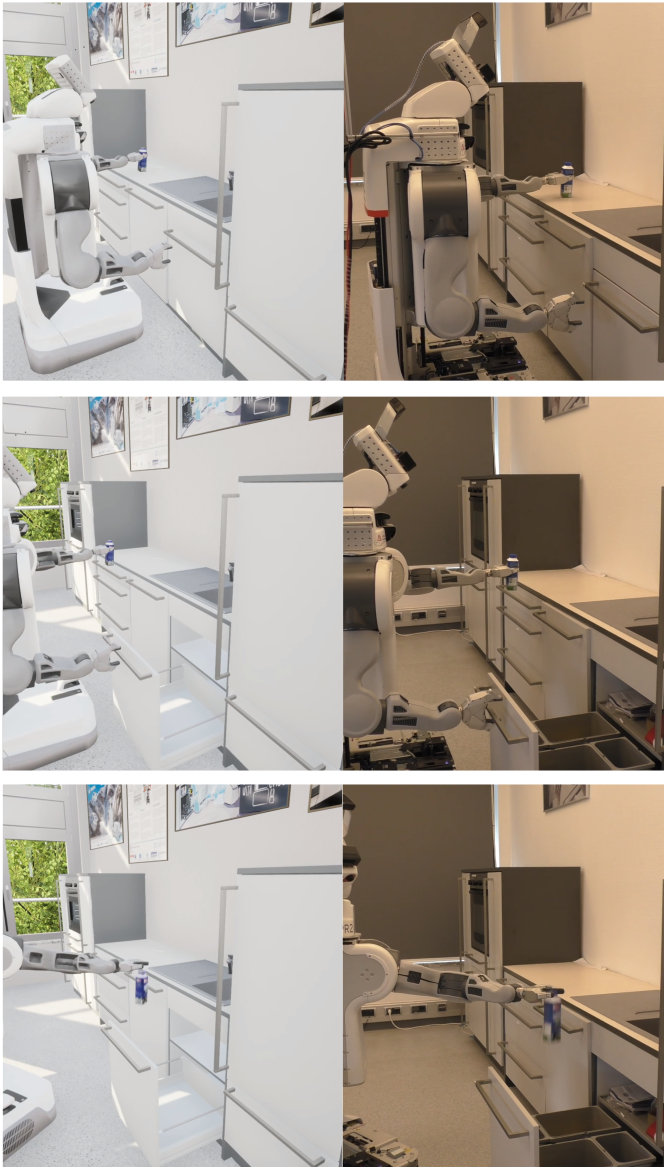}
  \caption{Belief-state (left) during action of real robot (right).}
  \label{fig:belief-state}
\end{figure}

In order to use \urobosim as belief state of the robot, we first mirror the joint state of the robot in simulation. If the real robot perceives an object it will be
added to the simulation. Objects grasped by the real robot will also be grasped in the simulation. Similar to the usage as mental simulation, we can use the belief state modus,
we can simulate the result of actions. If the real robot pulls open a drawer it is represented in the simulator. If an object is dropped, the simulator can provide a location
to search the object. The main difference is that it acts synchronously with the real robot. Figure \ref{fig:belief-state} shows the belief state during the execution of dropping milk into the trash bin inside a drawer. Physics is enabled during the whole simulation. This allows the robot to know for example, how far is the drawer open.

\section{Evaluation}
\label{sec:ex}
In this work, we want to evaluate two aspects of \urobosim.
The first aspect is the transfer from the real world perception-action loop to the simulated world.
The second aspect is that our simulator can be used to improve the robot's performance without including any real life data and allow even to learn correct models which cannot be learned from less realistic simulators such as \bulletworld.
To show that \urobosim~is able to cover those aspects, we selected the scenario where we want to use the same framework and modules as used in the real robot and let the robot learn to grasp the milk from the fridge door, as mention in Section \ref{sec:application-learning}.
To improve the robot's performance taking the milk from the fridge while using only the data generated from the simulator, \urobosim~requires to handle the same perception-action loop as the real robot.
This allows us to collect data to learn the required distributions.
Those distributions can be used later in mental simulation to plan the correct actions and let them afterward be executed by the real robot.

We performed the transportation of the milk 50 times in \urobosim. The acquired \neems~are available under the following link\cite{data}. 
From this data, we extracted the robot's x and y coordinates from which it grasped the milk successful from the fridge door.
The data is available under the following link\cite{data-transformed}.
Those points were used to create a multivariate Gaussian from which the robot can sample to determine a suitable grasping position.

\begin{figure}[htb]
	\centering
	\includegraphics[width=0.45\columnwidth,height=0.45\columnwidth]{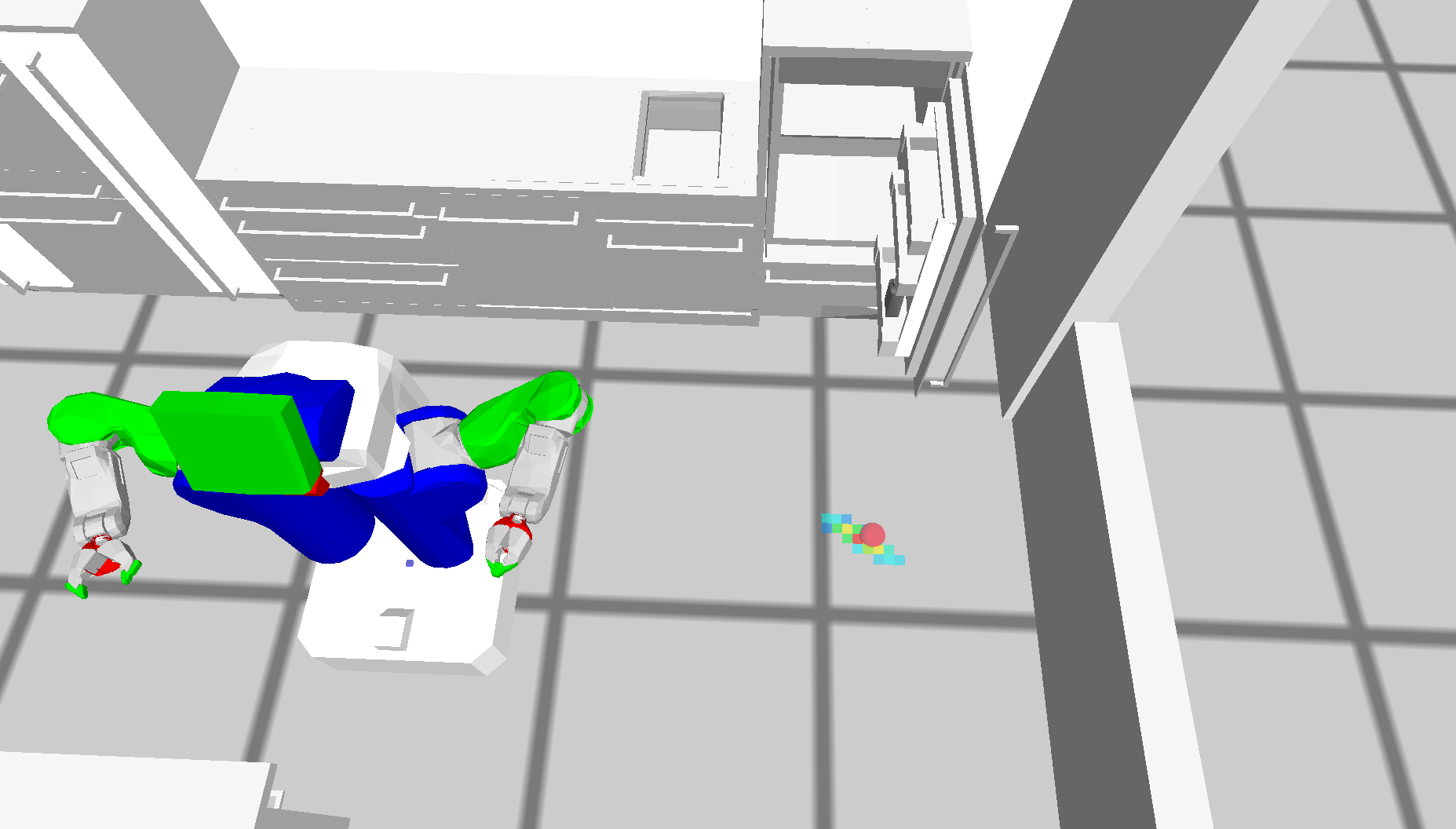}
	\includegraphics[width=0.45\columnwidth,height=0.45\columnwidth]{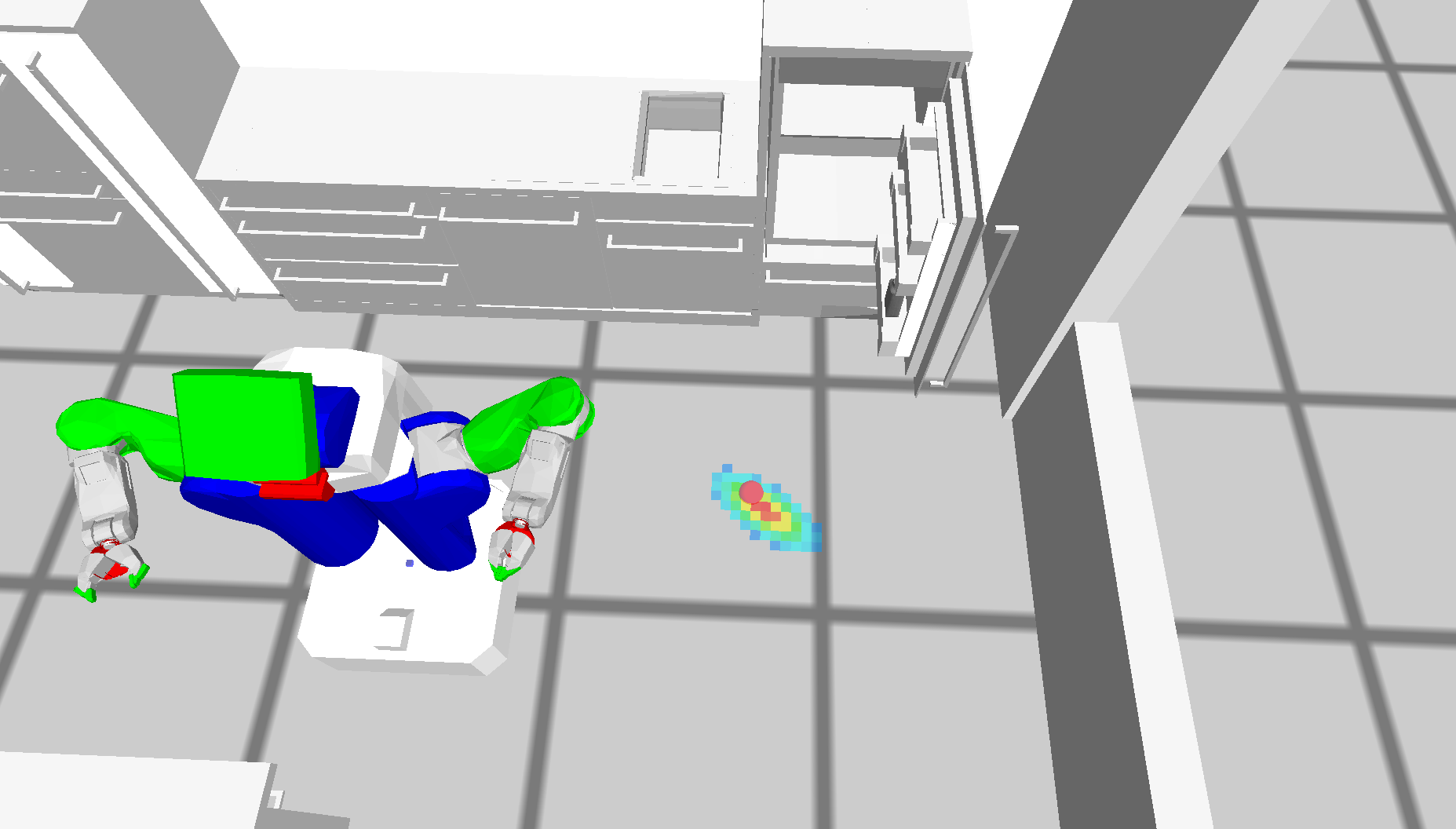}
	\caption{Learned distributions to grasp the milk from the bottom shelf fridge door: (left) distribution generate from \bulletworld, (right) distribution generate from \urobosim.}
	\label{fig:learning}
\end{figure}

Figure \ref{fig:learning} shows left the distribution created by the points from \bulletworld~and right the distributions created with \urobosim.
You can see that the distribution created from \bulletworld~forces the robot to position itself parallel to the edge of the door.
In real life, this position causes that the robot is not able to detect the milk and this leads to a failure.
The distribution from the right forces the robot to position itself towards the milk box so it can detect it better.

Using the newly created model, we managed to reduce the retries during the fetch and deliver action of the milk from an average 4.02, 4.43 standard deviation to 1.78 and 2.14 standard deviation during simulation.
W.hich led to an improvement of 55.72 percent.
Once we observed the performance improvement, we used the statistical models to let perform the action in mental simulation and afterward we observed that the robot was capable to execute the same action in real life.

One can argue that the chosen models for this evaluation cannot cope with changes in the environment. 
Developing robust and flexible models is not the goal of this paper and rather showing the functionality of \urobosim.
A detailed overview of how we are developing statistical models that cope with a dynamic world is shown in \cite{koralewski2019self}.



\section{Related Work and Discussion}
\label{sec:related}

Mental simulation plays an important role in interactions with the environment for humans. Peter W Battaglia et al. in \cite{battaglia2013simulation}
analyze the relation between simulations and the reasoning about physics. Inside a simulation, they created different
scenarios for which humans should predict for example if a tower will fall and in which direction will it fall. It is shown that
humans rely on rich probabilistic simulations and experience.
Inspired by this work, Wei Liang et al. in \cite{liang2015evaluating} evaluates the human cognition of
containing using simulations. They developed a simulator to test the
affordance of containers and compare it to the human prediction.

In recent years, the concept of mental simulations in perspective of cognitive science
was analyzed by Tomer D. Ullman in \cite{ullman2017mind}. They hypothesized that intuitive
physical inferences are based on mental simulations. Furthermore, the
provided possibilities to close the gap between mental simulation of
human and robot by using physic engines.

Robot simulator using physic engines exist in different forms with different
capabilities and applications. One of the most known ones is Gazebo. It is integrated
into ROS and provides multiple features but misses two features
required for our goal. First, it has no photorealistic renderer. The second feature we require is good scalability. We
aim for complex environments such as apartments and kitchen with many
objects. Others face similar problems, which results in a number of
new simulators using game engines such as Unreal Engine 4 or
Unity.

One such simulator focuses on UAVs and autonomous driving. Airsim is presented by
Shital Shah et al. \cite{shah2018airsim}. It provides the capabilities to
control, monitor and collect data from different UAVs. Based on this simulator
Elizabeth Bondi et al. in \cite{bondi2018airsim} developed Airsim-W. By building a
simulation environment for wildlife Conservation they want to generate labeled data as well as test algorithms
Furthermore they want to improve the detection and counting of animals as well as
poachers in wildlife conservatives. 

Another simulator developed  by  Pablo Martinez-Gonzalez  et al.,  is
UnrealRox \cite{martinez2018unrealrox}. The focus of this simulator is generating image data from
the perspective of robots. The  system allows to generate large amount
of data and increase the accuracy of perception tasks.


Filipe Figueredo Monteiro et al. in  \cite{monteiro2019simulating} give an introduction to Nvidea
Isaac. Nvidea Isaac is integrated into a fork of Unreal Engine 4. It
consists of a complete framework for machine learning such as object
recognition, image segmentation and path planning.

\section{Conclusion and Future Work}
\label{sec:conclusion}

In this work, we showed that we can perform real robot plans in our simulator under realistic conditions. Using this we were able to improve the learned model for robot positions for grasping milk from the fridge. While the number of retries could not be reduced to zero it could be reduced. We also managed to eliminate the occurring high-level errors.
The current experiment does not explore every physical capability of the simulation. While the continuous change of the fridge door or the placement of the milk uses physics, unreal engine 4 provides more complex physics such as fluids or deformable objects. But it is also a fact that the physical parameters such as viscosity of fluids are mostly unknown to the simulator, making it difficult to accurately simulate the result of actions such as pouring. Therefore we want to explore the possibility to learn the progress of the purring action. Depending on the speed of how fast the
glass is filling, we can learn the timing when to stop purring. We also want to explore the possibility to learn simulation parameters from real experiments in order to improve the accuracy of the simulation.


\section*{Acknowledgments}
\begin{small}
\noindent
This work was supported by DFG Collaborative Research Center \emph{Everyday Activity Science and Engineering (EASE)} (CRC \#1320) and the DFG Project \emph{PIPE} (project \#322037152).
\end{small}

\allbibliography{literature}
\bibliographystyle{IEEEtran}

\end{document}